%% file: main.tex
\documentclass[11pt]{article}
\usepackage{acl}

\usepackage{times}
\usepackage{latexsym}
\usepackage[T1]{fontenc}
\usepackage[utf8]{inputenc}

\usepackage{algorithm}
\usepackage{algpseudocode}
\usepackage{graphicx}
\usepackage{booktabs}
\usepackage{xcolor}
\usepackage{amsmath}

\title{MemCtrl: Using MLLMs as Active Memory Controllers on Embodied Agents \vspace{-0.5cm}}

\author{
Vishnu Sashank Dorbala \quad
Dinesh Manocha \\ \\
University of Maryland, College Park
}

\definecolor{darkgreen}{rgb}{0,0.5,0}

\begin{document}
\maketitle
\input{sec/0_abstract}    

\section{Introduction}
\label{sec:intro}

\begin{figure}
    \centering
    \includegraphics[width=\linewidth]{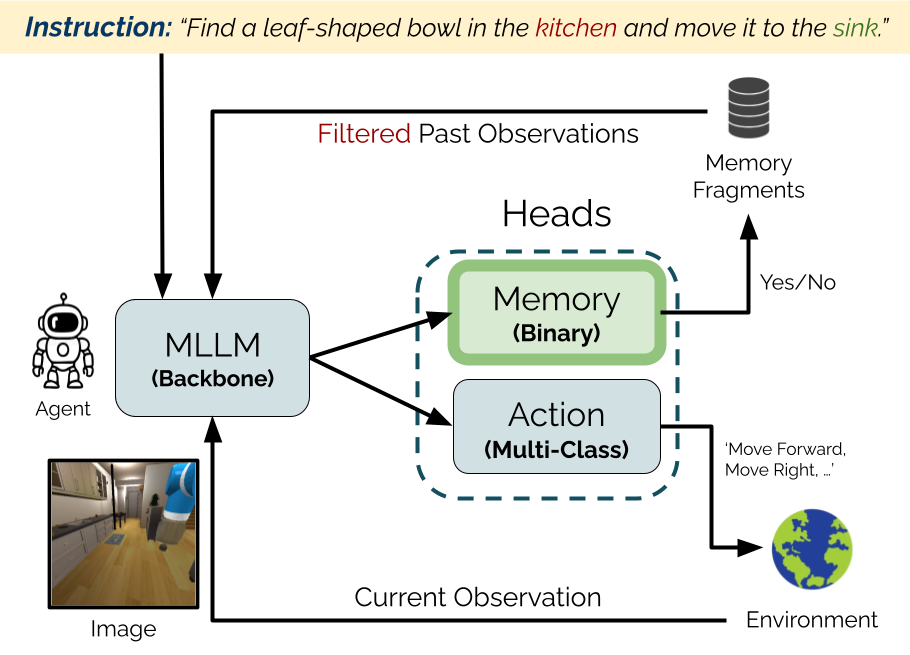}
    \caption{\textbf{Overview}: We present MemCtrl, a novel memory filterting scheme to improve decision making performance on \textbf{small} MLLMs tackling embodied tasks.
    %Unlike prior work involving a post-filtering of gathered observations for context (RAG), in this scheme, 
    Our approach proposes a \textit{trainable memory head} (green box labeled ``Memory'') that learns to actively filter out redundant observations on-the-go. This form of \textit{active} filtering alleviates issues with inefficient retrieval from stored observations, while also enabling scalability as a detachable memory head.
    }
    \label{fig:cover}
\end{figure}

An overarching goal of Embodied AI is the development of a generalist agent that can perform consistently well with high success on diverse tasks, environments and instructions \citep{szot2025multimodal}. A common paradigm to achieve this has been to utilize foundation models to develop task solving frameworks \cite{mu2023embodiedgpt, yang2025embodiedbench}. While a few of these methods generalize well to diverse tasks and instructions \citep{driess2023palme, zawalski2024robotic}, they are constrained by high training costs, prohibiting them from quickly being able to adapt to novel real-time settings where the data is out of distribution. Further, finetuning large foundation models incurs significant computational capacity, proving to be a significant hurdle in the democratization of these methods \citep{liang2022code}, especially in the context of robotics, where computation on the edge is of vital importance.

\begin{figure*}[t!]
    \centering
    \includegraphics[width=\linewidth]{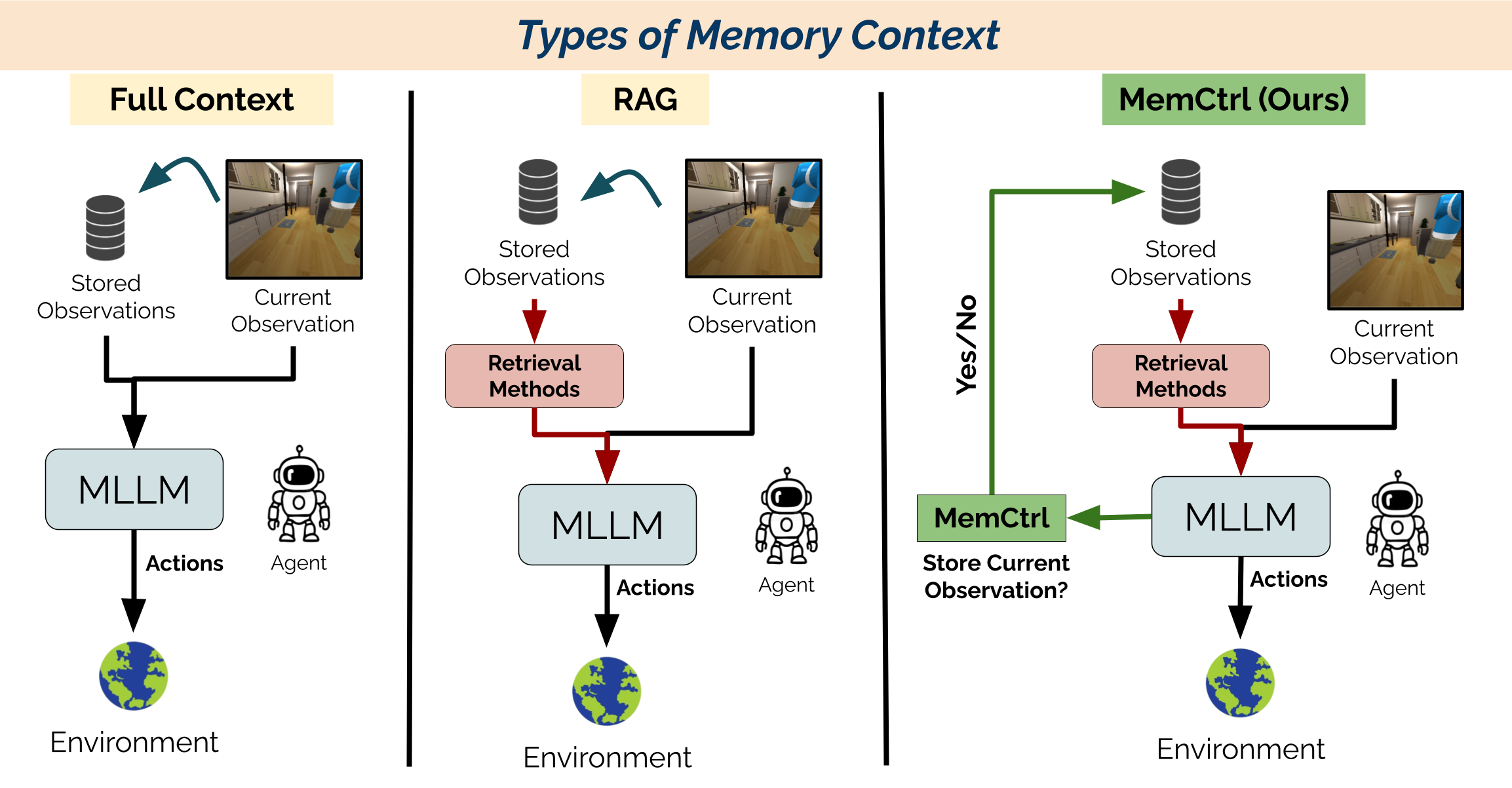}
    \caption{\textbf{Comparison with Prior Work}: We present MemCtrl, a novel approach to train \textit{``memory heads''} to filter observations on the go. Prior work either used the entirety of stored observations as context (left) or filtered them via a variety of Retrieval Augmented Generation (RAG) based schemes ({\color{red} red} arrows), both of which assume the parsing of large amounts of data offline. MemCtrl introduces transferrable heads to use on MLLM backbone ({\color{olive} green} arrows) to actively filter observations.}
    \label{fig:comp}
\end{figure*}

An alternative, more feasible paradigm has been a modular system where foundation models are used in conjunction with memory banks \citep{zhong2023memorybank, wang2024memoryllm} of past experiences and reflections. Foundation models including very large Multimodal Large Language Models (MLLMs) such as LLaMA 4 \cite{llama} and Deepseek V3 \cite{deepseek}
are limited by the size of their context window, and developing methods to refine and selectively pass memory as context is an active area of research \citep{wu2025human}. Prior work in this area includes use of intrinsic model editing techniques for memory injection~\citep{mitchell2022mend, meng2023memit}, or extrinsic interactions with episodic logs, Retrieval-Augmented Generated (RAG) \cite{gao2023retrieval}, or long-range latent states~\citep{park2023generative, wang2023voyager}.  

While both paradigms have shown improved performance in multi-step reasoning, their implementation on embodied robot agents raises practical issues. Embodied agents providing assistance often use small models ($<20$B parameters) that work locally on-device, with often limited or only cloud access to large memory storage. Moreover, these agents need to generalize to novel settings, making it preferable to have modular, lightweight segments that are easily \textit{transferrable}. 

% A few works explore using a foundation model for \textit{actively} writing memory, but these are usually limited to summarization tools like Voyager \citep{wang2023voyager} and Reflexion \citep{shinn2023reflexion} that create summaries aimed to explain or reason about past observations before decision making. This reasoning however tends to be unstructured and uncontrolled.

To model a more efficient memory framework for embodied agents, we draw inspiration from how humans store memories of experiences. While performing various embodied tasks, humans do not accumulate every observation for later retrieval, but rather learn to actively filter out only certain vital fragments that we assume to be relevant to our task \cite{eswm}. When queried, we reconstruct the missing fragments of memory through commonsense reasoning. This makes us humans highly efficient reasoners even with limited storage.

We aim to endow compact embodied agents with a similar ability: rather than relying on large external memory banks or complex retrieval pipelines, the agent must actively learn to store vital memories while filtering redundant ones on the go. Learning this skill across a wide range of tasks would enable scalable self-improvement under tight computational and memory budgets.

\noindent{\bf Main Results:} 
To address these issues, we present \textit{MemCtrl}, a transferrable memory augmentation scheme that aims to improve the embodied decision making performance of \textbf{small} models.
MemCtrl introduces a trainable memory head that learns to selectively store \textit{memories of importance}, increasing both parameter and memory efficiency for self-improving embodied agents.
Our contributions are as follows:
\begin{itemize}
    \item \textbf{Active Memory Filtering:} 
    We introduce two lightweight memory heads $\mu$ trained on top of a frozen MLLM backbone to actively filter observations to determine which to keep and which to discard in memory. Unlike prior retrieval-based work involving filtering large observational data offline, $\mu$ enables the MLLM to engage in real-time filtering, which is particularly useful in memory-constrained settings involving small models. 

    \item \textbf{Transferrable Heads:}
    $\mu$ is model-agnostic and attaches to any off-the-shelf MLLM without having to finetune or edit the backbone to remove redundant observations for more prudent decision making. This modular design allows MemCtrl to transfer across embodied setups and vision-language backbones, enabling its scalable transfer across embodied agents in diverse settings.  

    \item \textbf{Improving Small Model Performance:}
    Finally, attaching $\mu$ to the worst performing agents on the Habitat \cite{puig2023habitat} and ALFRED \cite{shridhar2020alfred} splits of EmbodiedBench \cite{yang2025embodiedbench} shows a significant improvement on task performance of around $16\%$ on average, while also storing significantly fewer observations. This efficiency makes MemCtrl favorable for real-world deployment.
\end{itemize}

\section{Related Work}\label{sec:related_works}

\subsection{MLLMs in Embodied AI}
Several works in recent literature have leveraged large language models (LLMs) for high-level robot planning. For example, \citet{saycan2022arxiv} introduce a framework (SayCan) where an LLM translates natural-language instructions into feasible robot actions constrained by a set of learned skills. This approach demonstrated that LLMs can provide semantic task knowledge, but it relies on a fixed library of affordance-grounded skills and struggles with adapting to novel situations. Recent multimodal LLMs extend this idea by directly integrating visual inputs. For instance, PaLM-E \citep{driess2023palme} is a vision-language model that outputs robotic actions, achieving good generalizability across manipulation and navigation tasks. However, PaLM-E requires a lot of training data, limiting its real-time adaptability. Similarly, RT-2 \citep{zitkovich2023rt2} augments a vision-language model with web-scale pretraining to create a vision-language-action (VLA) agent for improved zero-shot object understanding. However, it makes use of short temporal contexts without an explicit memory mechanism. In contrast, our work uses an MLLM as an active memory controller, enabling the agent to retain and recall cross-modal information over long horizons. Our design empowers embodied agents with small models to handle complex, extended tasks without requiring large-scale training.

\subsection{Memory-Augmented Agents}
Recent works have explored augmenting LLM-based embodied agents with memory to address challenges with long-horizon task solving. \citet{mai2023llmbrain} propose using an LLM itself as a `robotic brain' that maintains an egocentric memory of the agent’s observations and dialogue. Their system shows that a textual memory stream can help the agent refer back to important context, improving consistency in multi-step tasks. Another approach is to attach an external memory to the agent. In the HELPER framework \citep{sarch-etal-2023-open}, they  maintain a repository of dialogue-to-action examples, retrieving relevant past interactions to condition the LLM when parsing new instructions. 
They show that dynamic memory of prior events or user preferences can overcome the limitations of fixed prompts or short context windows to improve task completion success. However, this still relies on a separate module for populating memory, with the foundation model having a passive role. In contrast, our work introduces a framework that enables the MLLM to \textit{actively} control what and what not to store in the agent's memory. Figure \ref{fig:comp} highlights this idea. The MemCtrl module takes in the embedding of the current observation from the MLLM and determines whether or not the observation should be stored.

% \subsection{Generative World Models}
% Recent work has explored the use of generative world models to imagine future states of subgoals. \citet{black2023susie} propose using a pretrained image-editing diffusion model as a high-level planner. Their system (SuSIE) generates hypothetical future observations—visual subgoal images—given the current camera view and a command, enabling zero-shot manipulation planning for previously unseen objects. In a similar vein, DINO-WM \citep{zhou2024dinowm} learns a latent world model on top of pre-trained visual features and demonstrates zero-shot plan generation in new environments without task-specific training. Such generative approaches leverage rich prior knowledge (from internet-scale image datasets or self-supervised learning) to guide robots when explicit environment models are unavailable. In our work, we use generative world modeling as a forecasting tool that the MLLM may choose to use.

\section{MLLM-based Embodied Agents}\label{sec:prelim}

Let $\mathcal{M}$ be an MLLM, $\mathcal{O}_{c}$ be the current observation, and $\mathcal{I}$ be the instruction. A baseline method utilizes $\mathcal{M}$ to translate the observations and instructions into actionable outputs for the agent as follows:
\[
a = \mathcal{M}(\mathcal{O}_{c}, \mathcal{I}) ,
\]

\noindent where $a \in \mathcal{A}$ such that $\mathcal{A}$ represents a set of feasible actions for the agent in the environment.
Using $\mathcal{M}$ as a prior in this ``zero-shot'' manner leads to subpar performance, since the agent does not have any continual context of the environment, and takes actions solely based off of the current image observed and its capacity for commonsense reasoning \cite{clipnav, majumdar2022zson, cow, shah2023vint}.

\subsection{Memory-Augmented MLLM Agents}

One way to improve zero-shot performance of $\mathcal{M}$ is to provide continued context to the agent as a set of past observations, i.e., $\{\mathcal{O}_{c}, \mathcal{I}, \mathcal{C}\}$, where $\mathcal{C}$ is the context passed to $\mathcal{M}$:
\[
\mathcal{C} = \{R_{i}, \mathcal{O}_{i}\}_{i: 1 \rightarrow n}.
\]
$R_{i}$ and $\mathcal{O}_{i}$ respectively represent the $i^{th}$ reflection and observation in $n$ timesteps.

EmbodiedBench \cite{yang2025embodiedbench} highlights that adding history as context greatly influences performance, particularly when it comes to embodied tasks involving long-horizon instructions.
Further, they highlight benefits to adding memory in the form of agent reflections, where the agent reflects on its past interactions with the environment to refine its future plan.

While adding history improves zero-shot performance, $\mathcal{M}$ is limited by a context window $h$, and $n < h$. To circumvent this issue, several approaches explore retrieval pipelines to compress memory to obtain context $c = F(\mathcal{C}, \mathcal{I})$ for $\mathcal{M}$. $F$ here is retrieval function that selects the most relevant parts of the whole context given an objective, which in this case is the instruction $\mathcal{I}$.

\noindent \textbf{Inefficient Retrieval}:
In alignment to other reported results \cite{lostmiddle, packer2023memgpt}, we similarly observe that as the size of context $n = sizeof(\mathcal{C})$ increases it leads to more inefficient retrieval, especially since we are limited by a context window of size $h << n$. Robot agents running in the wild often collect observations at high frequencies ($>1$ observation per second), which quickly increases the size of $n$.
Further, having redundant observations in memory only adds to this issue, prompting the development of better strategies for write-time memory control, especially on small models.

To achieve such active memory control, we propose \textit{MemCtrl}, a learnable memory augmentation scheme that allows active write-time memory control on the model $\mathcal{M}$. We describe this in the following section.

% In a case of \textit{complete memory}, $n$ is treated as the upper bound and $i$ is sequential with a difference of $1$. This means that every single context vector $\mathcal{c} \in \mathcal{C}$ where $c = [r, o]$ are stored upto $n$ spaces to be passed onto $\mathcal{M}$.

% While passing complete memory provides more dense information, this method becomes intractable after a few episodes, since all MLLMs $\mathcal{M}$ have limited context windows. Smaller models like also tend to have smaller context windows, which explains their poor performance. A common workaround for this is \textit{selective memory}, or having to selectively choose relevant context from $\mathcal{C}$ to pass to $\mathcal{M}$. However, this selection is usually done via external compression mechanisms, and the MLLM itself has little to no say on what memory to store.

\section{MemCtrl: Training Memory Heads ($\mu$)}\label{sec:method}

\begin{figure}
    \centering
\includegraphics[width=\linewidth]{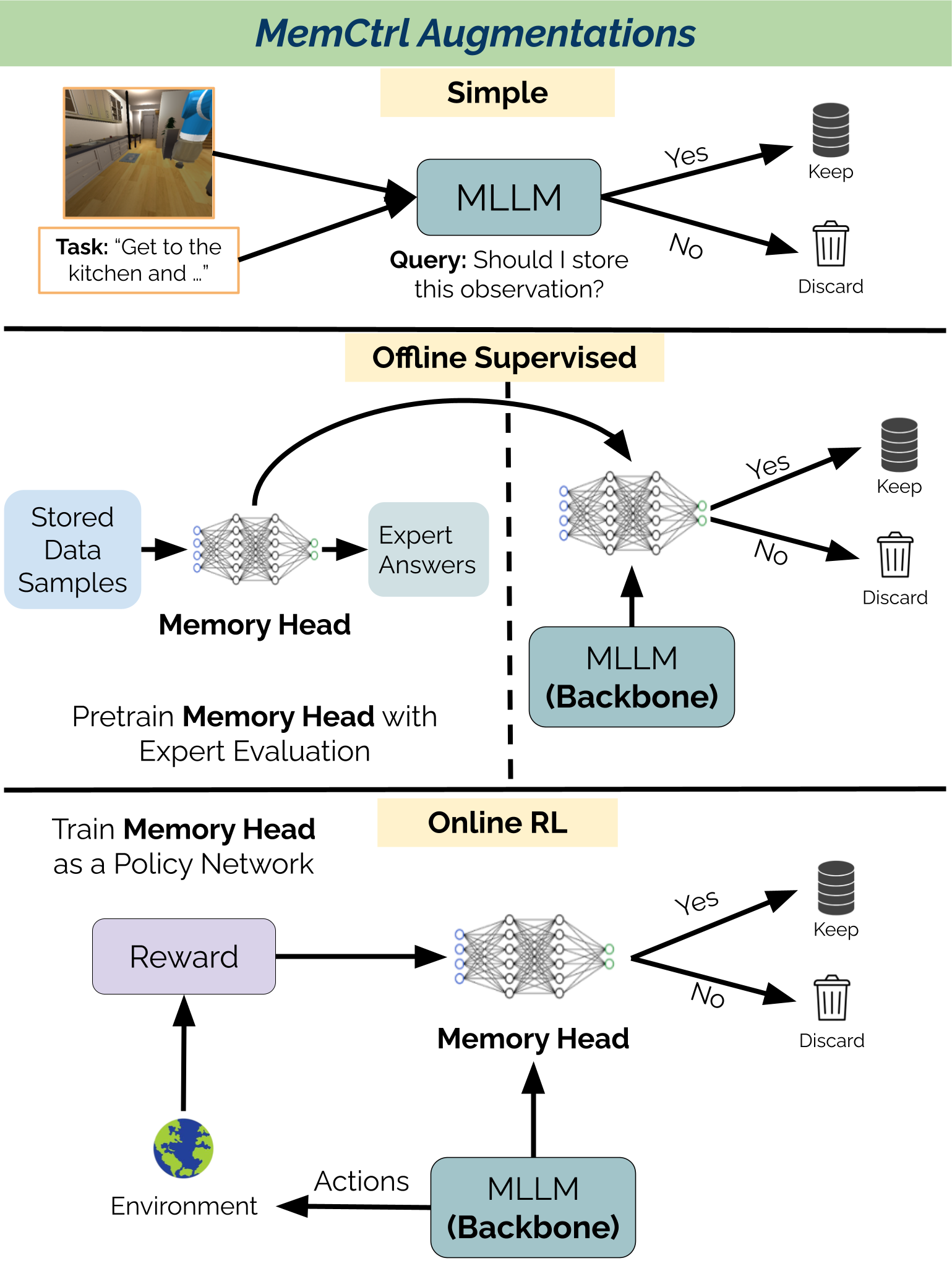}
    \caption{\textbf{MemCtrl}: We experiment with $3$ augmentations. The simple case acts as a non-trained baseline, where the MLLM is directly queried about storage. In the offline supervised case, $\mu$ is first pretrained using expert answers from a high performing, expert MLLM (GPT-4o here). This trained binary classifier then acts as a head on top of the MLLM backbone. In the Online RL case, we train the memory head online as a policy network. We use a sparse reward on task success and a dense reward on action success. Note that MemCtrl is trained as a detachable head that takes the visuolingual MLLM embeddings as input.}
    \label{fig:approach}
\end{figure}

% WHAT IS THE GOAL OF "TRAINING MEMORY HEAD"? CAN YOU SPECIFY THIS GOAL USING MATHEMATICAL EQUATIONS? ARE YOU TRYING TO OPTIMIZE SOME FUNCTION?
To achieve write-time memory control, we consider three natural augmentations to $\mathcal{M}$, illustrated in Figure \ref{fig:approach}. Our objective is to improve decision making on \textbf{small} visual-language models, which also translates to small context windows $\mathcal{C}$'s, by empowering them to better filter observations prior to storing them in memory.

% Integrating memory with foundation models has been essential for overcoming the challenge of limited context windows. A popular way to do this has been with RAG \cite{gao2023retrieval}, where the context $\mathcal{C}$ is determined from a retrieval policy on stored data. In the context of robotics, this stored data is often described as multi-modal observations of the scene, which is then parsed to choose information relevant to the instruction. However, a key observation here is that the stored data is \textit{offline} and is usually collected and updated during exploration. As the agent explores more space, it gathers more observations, requiring increasingly more sophisticated RAG pipelines for compressive memory retrieval for $\mathcal{C}$.

% We propose three  involving \textit{active} agents with small models.
Filtering observations in this way fully avoids the problem of inefficient retrieval described in the previous section, since the agent's context is now driven by its own decisions, much akin to humans deciding only to remember moments of importance that might be meaningful to them in the future.

For this, we propose a \textbf{trainable memory head $\mu$} as a simple binary classifier that learns to either keep or discard memory. $\mu$ integrates with the $\mathcal{M}$ backbone, to make decisions online about whether or not to store the current observation.
Following on the our definitions so far, we get
\begin{equation}
    \begin{aligned}
    c &= F(\mathcal{C}, \mathcal{I}), \\
    a &= \mathcal{M}_{a}(\mathcal{O}_{c}, \mathcal{I}, c), \\
    b &= \mathcal{M}_{\mu}(\mathcal{O}_{c}, \mathcal{I}, c).
    \end{aligned}
\end{equation}
\noindent where $b \in \{0, 1\}$ is a binary classifier that determines if the current observation must be added or discarded. The updated context, $C'$ can then be written as,

\begin{equation}
      C' =
  \begin{cases}
    C \ \cup \ \{(\mathcal{O}_c, a)\} & b = 1, \\
    C & \text{otherwise.} \\
  \end{cases}
\end{equation}

% $\mu$ empowers $\mathcal{M}$ for active memory control, allowing it to directly filter the incoming observations. 

\noindent We consider $2$ ways to integrate $\mu$ onto $\mathcal{M}$:

\begin{itemize}

    \item \textbf{Offline, Fully-Supervised}: We first gather offline data from a high-performing $\mathcal{M}$ treated as an expert. Using the gathered negative and positive samples, we train $\mu$ offline as a binary classifier to determine which observations led to success and which led to failure. We transfer this \textit{pretrained} memory head onto a low-performing model $\mathcal{M}$, as an expert supervision gate. We train the network with a binary cross-entropy loss function:
    \begin{equation}\label{eq:ce}
        L(y, \hat{p}) = y\log(\hat{p}) + (1-y)\log(1-\hat{p}),
    \end{equation}
    where $y_i \in [0,1]$ is the ground-truth label and $\hat{p}_i$ is the predicted probability from $\mu$ of whether the current observation should be stored.
    \item \textbf{Online RL}: We directly train the memory head and the action head via an online RL policy. We model two rewards: 1) a sparse reward for episode success, and 2) a dense reward for picking valid actions:
    \begin{equation}\label{eq:reward}
        R(r, a) = r + \mathbf{1}_{a \in \mathcal{A}},
    \end{equation}
    where $r\in \{0,1\}$ is the binary reward signal for task completion, and $\mathbf{1}$ is an indicator function. In our approach, $r=0$ for all steps except for goal-completing ones. These reward functions ensure that $\mu$ picks helpful observations and the action heads make valid decisions.
    % In this setting, the memory head is trained online, and we aim to improve performance, allowing $\mu$ to selectively control the context $\mathcal{C}$ passed to $\mathcal{M}$. 
\end{itemize}

% We additionally implement a simple baseline model where the MLLM directly predicts if to store the observation or not.
\noindent In both these cases, the memory head $\mu$ empowers the MLLM to play an active role in filtering observations, as highlighted in Figure \ref{fig:approach}. Further, $\mu$ is a head, and can be transferred across arbitrary MLLMs helping alleviate the cost of directly finetuning MLLMs. Algorithm \ref{alg:supervised_method} and \ref{alg:rl_method} in the Appendix present the details of our algorithm for the supervised and RL variants, respectively.

\begin{table*}[t!]
\centering
\resizebox{\linewidth}{!}{
\begin{tabular}{l|cccccc|cccccc}
\toprule
\textbf{Model} &
\multicolumn{6}{c|}{\textbf{EB-ALFRED}} &
\multicolumn{6}{c}{\textbf{EB-Habitat}} \\
 & \textbf{Avg} & \textbf{Base} & \textbf{Common} & \textbf{Complex} &
   \textbf{Spatial} & \textbf{Long} &
   \textbf{Avg} & \textbf{Base} & \textbf{Common} & \textbf{Complex} &
   \textbf{Spatial} & \textbf{Long} \\
\midrule
Gemma-3-12B-IT & $25.6$ & $32$ & $26$ & $38$ & $20$ & $12$ & $23.0$ & $58$ & $10$ & $24$ & $24$ & $4$ \\
Gemma-3-12B-IT + $\mu_{\textbf{Simple}}$ & $28$ & $41$ & $27$ & $34$ & $16$ & $22$ & $31.2$ & $62$ & $\underline{15}$ & $30$ & $31$ & $18$ \\
Gemma-3-12B-IT + $\mu_{\textbf{Offline Sup.}}$ & $\textbf{32.2}$ & $\underline{48}$ & $\underline{31}$ & $32$ & $\underline{23}$ & $\underline{27}$ & $31.8$ & $\underline{66}$ & $14$ & $35$ & $27$ & $17$ \\
Gemma-3-12B-IT + $\mu_{\textbf{Online RL}}$ & $27.8$ & $38$ & $29$ & $\underline{41}$ & $21$ & $26$ & $\textbf{33.8}$ & $60$ & $13$ & $\underline{37}$ & $\underline{35}$ & $\underline{24}$ \\
\midrule
Qwen2.5-VL-7B-Ins & $4.7$ & $10$ & $8$ & $6$ & $0$ & $2$ & $14.3$ & $32$ & $2$ & $26$ & $14$ & $2$ \\
Qwen2.5-VL-7B-Ins + $\mu_{\textbf{Simple}}$ & $9.6$ & $\underline{15}$ & $10$ & $12$ & $2$ & $14$ & $21.4$ & $\underline{39}$ & $\underline{10}$ & $31$ & $14$ & $13$ \\
Qwen2.5-VL-7B-Ins + $\mu_{\textbf{Offline Sup.}}$ & $12.2$ & $9$ & $10$ & $14$ & $2$ & $\underline{26}$ & $\textbf{22.8}$ & $\underline{39}$ & $5$ & $33$ & $\underline{17}$ & $\underline{20}$ \\
Qwen2.5-VL-7B-Ins + $\mu_{\textbf{Online RL}}$ & $\textbf{14.2}$ & $10$ & $\underline{13}$ & $\underline{21}$ & $\underline{3}$ & $24$ & $22.2$ & $37$ & $3$ & $\underline{37}$ & $16$ & $18$ \\
\bottomrule
\end{tabular}
}
\caption{\textbf{Results}: We augment \textit{Qwen2.5-VL-7B-Ins} and \textit{Gemma-3-12B-IT} with the 3 variations of MemCtrl on 5 subsets of EB-ALFRED and EB-Habitat \cite{yang2025embodiedbench}. Note the improve performance overall of adding the memory head $\mu$. In particular, we note superior performance on long context and complex instructions, which tend to be long horizon where memory helps. Performance on Habitat is also much better than ALFRED overall, hinting that navigation heavy tasks that are common on the Habiat dataset might benefit from memory augmentation more than manipulation ones common in ALFRED.}
\label{tab:main_results}
\end{table*}

\section{Experimental Setup}\label{sec:exp_setup}

\paragraph{Datasets.} Our objective is to improve the performance of an MLLM by augmenting it with a memory head. For this, we choose EmbodiedBench \cite{yang2025embodiedbench} as a benchmark for evaluation, since it provides us with tools to automate embodied task evaluation on both ALFRED \cite{shridhar2020alfred} and Habitat \cite{puig2023habitat} simulators, making it easy to evaluate the performance of multiple MLLMs on various tasks. We modify this evaluator with memory heads for our task.

\paragraph{LLM backbones.} To showcase improvement, we choose two low-performing models on EmbodiedBench, \textit{Qwen2-VL-7B-Ins} and \textit{Gemma-3-12B-IT} aiming to showcase an improvement in performance when augmented with MemCtrl. We run all models locally on a NVIDIA A5000 GPU.

\paragraph{Baseline: Simple, In-Context Learning.} We prompt $\mathcal{M}$ for a binary output on whether or not to store the current observation in its memory. We do not train a memory head here, but ablate with combinations of $\mu$ and $\mathcal{M}$. 

\noindent\textbf{Memory Heads.} Both $\mu$'s are paramterized as Linear MLP's with 3 layers, that map the backbone MLLM's embedding to a binary output. 

For the offline, supervised $\mu$, we first gather expert data using a high performing MLLM, GPT-4o which gives us a set of $X = [x_{1}, x_{2}, \dots, x_{n}]$ embeddings per episode mapped to a binary episode success or failure $l$, giving us $[n, 1]$ training pairs for each episode. We ensure balancing of the dataset with negative and positive samples and then train the MLP to overfit using a cross-entropy loss.

For the online $\mu$, we similarly define an MLP as the policy network that predicts a binary outcome. We define a sparse and a dense reward as described in the previous section, and train using REINFORCE \cite{reinforce}.

\section{Results}\label{sec:results}
Table \ref{tab:main_results} presents the main results, showcasing the benefits of MemCtrl across two different LLM backbones, \textit{Gemma-3-12B-IT} and \textit{Qwen2.5-VL-7B-Ins}. We pick these two models as they perform among the worst on the EmbodiedBench benchmark, aiming to show improved performance with the inclusion of our memory head.

\noindent\textbf{Increased Performance with $\mu$.} The overall performance improves across EB-Alfred and EB-Habitat with adding any type of memory head $\mu$. This is expected, since any form of memory provides continual context that is more meaningful for the MLLM.

In particular, we observe huge improvements on Long instructions, with results on \textit{Gemma-3-12B-IT} bumping from $12$ on the baseline to $26$ with the $\mu_{RL}$ augmentation, and \textit{Qwen2.5-VL-7B-Ins} going from $2$ to $24$. We believe this to be a result of a more strategic memory storage needed for longer-horizon tasks, like with EB-Habitat.

\begin{figure*}[t]
    \centering
    \includegraphics[width=\linewidth]{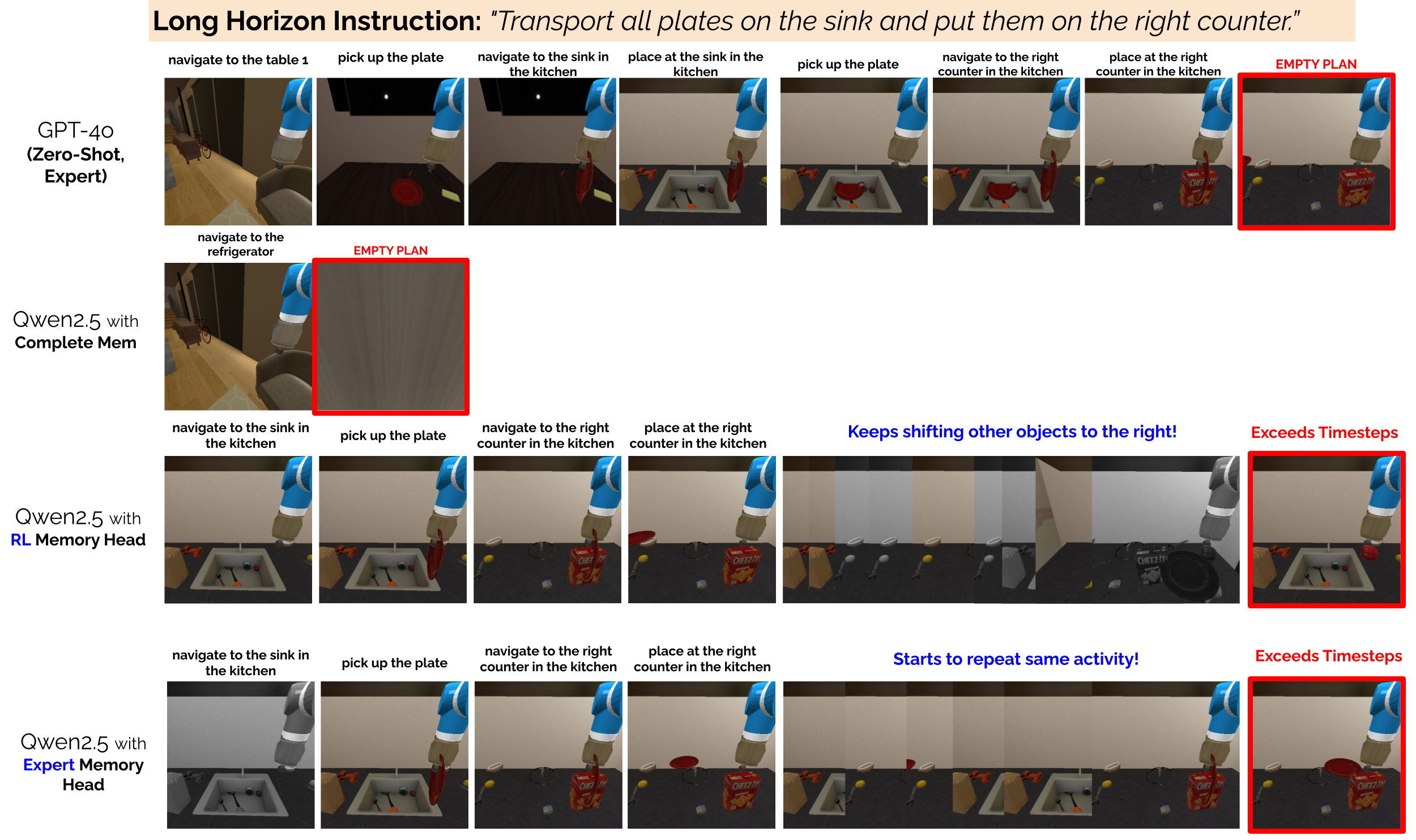}
    \caption{\textbf{Long Horizon performance on EB-Habitat}: We notice that on long horizon tasks, the expert tends to end the task early by hastily assuming that it is done (finishing after placing \textit{one} plate instead of \textit{all} plates). Memory heads highlight unique performance improvements, with $\mu_{\text{RL}}$ exhibiting a more \textbf{exploratory} nature by continuing to place \textit{new} objects at the right counter, and $\mu_{\text{Exp.}}$ being more \textbf{exploitative} by repeating the same activity over and over, with a single plate. Note: Grayed out images indicate discarded memories.}
    \label{fig:long}
\end{figure*}

We also observe an overall improvement in the task performance on complex instructions, where the instructions are not just long, but also contain irrelevant information. This is very evident with the Qwen model, where it goes up \textit{$3$}x from $6$ to $21$. For instance, the following is the difference between a base and complex instruction:- 
\begin{quote}
    \textbf{Base}: ``Move one of the pear items to the indicated sofa.''
\end{quote}

\begin{quote}
    \textbf{Complex}: ``When you find the fridge door open, go ahead and move an bowl to the sofa; otherwise, transport an hammer to the sofa.''
\end{quote}

The agent is expected to finish these tasks in a fixed set of timesteps, and over time, gathers more and more information about the environment as potential context for determining its next action. The base query here is fairly simple, requiring to track just a single object (`pear'). In contrast, the complex query not only has multiple objects to track (`fridge, sofa, bowl, hammer'), but is also sophisticated in its framing, requiring better reasoning. While more context would help with better reasoning, it also leads to more redundant information storage, which a trained memory head can help actively filter.

\paragraph{Performance across Gemma-3 and Qwen2.5.}
The baseline performance of Qwen2.5 as reported by EmbodiedBench is far lower than Gemma-3. We note much higher performance gains on Qwen2.5 compared to Gemma-3. Being one of the worst performing models on EmbodiedBench due to it's small size, adding a lightweight $\mu$ for active memory control bumps its performance up. Qwen2.5-VL + $\mu_{RL}$ is comparable to Ovis2-16B, a model with over twice the number of parameters that had $~16\%$ on the original benchmark.

\noindent\textbf{Alfred vs Habitat}:
Finally, we also notice better performance on Habitat overall compared to Alfred. Tasks in habitat tend to be more navigation centric, requiring more long-horizon planning. In contrast, Alfred focuses more on reasoning on current observations for manipulating objects. We infer that memory filtration is potentially more impactful on long-horizon navigation tasks, since it helps reduce redundant frames gathered.

\section{Qualitative Analysis}\label{sec:ablation}
% Table \ref{tab:memdiff} presents the percentage of memory stored by each method on average across all episodes. 
Figure \ref{fig:comparison} in the Appendix shows a qualitative example of MemCtrl in action. 

% \citet{eswm} recently studied fragmented memory as a world model, under the assumption that humans are capable of visually reconstructing the world and pathways in them simply by relying on memory fagments. The objective is to discover a function approximation $F$ that predicts the complete set of transitions given the partial memory and transitions, i.e., $q = F(\mathcal{C}, q^{*})$. In a similar vein, since the $\mu$ conditions the MLLM output to store fragments, we aim to analyze their \textit{reconstruction capacity} as a world model. We provide more details regarding these experiments in the supplementary.

\subsection{Visualization}

\begin{figure*}[t]
    \centering
    \includegraphics[width=\linewidth]{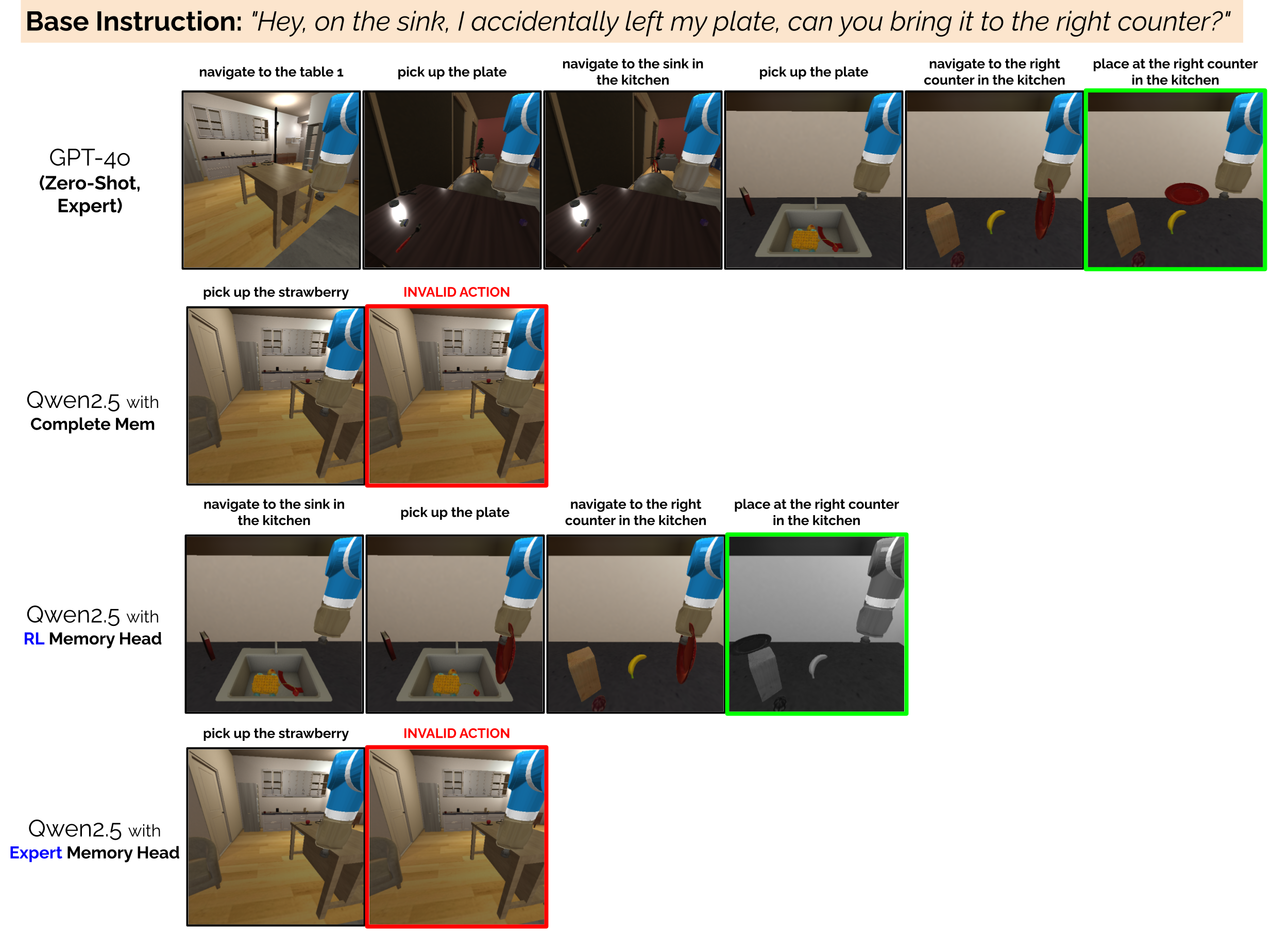}
    \caption{\textbf{Base Performance on EB-Habitat}: Here, we compare the performance of GPT-4o vs Qwen2.5-VL-7B-Ins, with various memory augmentations. While GPT-4V gives superior zero-shot performance, it is a very large model that is not easily finetunable. In this instance it also takes more steps to complete the task. On the other hand, $\mu$ boosts the performance of a significantly weaker model, and in this scenario, even doing it quicker with $\mu_{RL}$. The expert head fails here however, similar to the complete memory, causing the episode to end early. Note: Grayed out images indicate discarded memories.}
    \label{fig:base}
\end{figure*}

Figure \ref{fig:base} and Figure \ref{fig:long} compare a base and long episode, using GPT-4o and Qwen with and without memory. Note Qwen performs poorly on the EB-Habitat baseline, with only $14.3\%$ average vs GPT-4o that has $59\%$. Augmenting Qwen with a lightweight memory head $\mu$ bumps this up to around $22\%$, both in the case of $\mu_{RL}$ and $\mu_{Expert}$. 

In the base case, note the good zero shot performance of the GPT-4o agent in being able to complete the task. The complete memory agent however shows an invalid action after the first step, and this is also seen with the case of Qwen + $\mu_{Expert}$. However, with Qwen + $\mu_{RL}$ note the improved performance in being able to complete the task in fewer number of steps. Qwen is a much smaller model than GPT-4o, and with a small memory head augmentation, it is able to perform at par with its GPT-4o counterpart.

In the long horizon case, we make a few interesting observations. First, none of the $4$ methods seem to be able to successfully complete the task. The instruction requires \textit{all} plates to be transported to the right corner, requiring long-horizon task planning to keep track of which \textit{instances} of plates have been moved.
\begin{itemize}
    \item The \textbf{zero-shot expert} assumes that the task has ended after transferring the first plate, and hence sends no executable plan after being done. This stops the episode earlier than expected, hence failing the task.
    \item In the \textbf{complete memory case}, the agent is overloaded with too many experiences from the past in its memory, and this floods the MLLM with too much context, leading to a bad action being selected (\textit{navigate to the refrigerator}).
    \item In the \textbf{Qwen + $\mu_{\text{RL}}$} case, the agent successfully places \textbf{one} plate from the sink onto the right counter. While not ending the episode early like with the expert, it continues to place other objects including apples and wrenches in the vicinity to the right, ultimately running out of timesteps. This behavior can be associated with more active \textit{exploration}, as the agent seeks to explore new directions to complete the instruction. Note that the memory head filters out most of these extraneous observations (images in gray).
    \item In the \textbf{Qwen + $\mu_{\text{Expert}}$} case, we observe the opposite behaviour or \textit{exploitation}, where the agent just starts to repeat the same activity over and over again. It navigates to the sink, picks up a plate, transfers it to the right counter, then goes back to the sink, and so on. This form of exploitative behavior seems desirable, especially since the agent is tasked with transporting all plates to the sink. The expert memory head as a result decides to keep almost all of the observations, as the agent keeps doing the right thing in following the instruction. However, as the \textbf{same plate} is constantly being moved, it highlights a limitation of the expert memory head in being too conservative with classifying significant memories. 
\end{itemize}

\section{Conclusions}\label{sec:conclusion}

In this work, we present MemCtrl, a lightweight, transferrable memory framework that introduces a memory head $\mu$ on a backbone MLLM to actively filter \textit{memories of importance}. Instead of editing the MLLM directly or using external retrieval methods like RAG, $\mu$ is trained as a binary classifier decide whether or not to store current observations on the go. We introduce two ways to train $\mu$, and present a qualitative and quantitative analysis of $\mu$-augmented low parameter Qwen and Gemma models. Our results show significant performance improvement of around ~$16\%$ on average, across ALFRED and Habitat splits of the EmbodiedBench dataset. Further, we note the superior performance on instructions involving complex or long-horizon language.
% Our mechanism takes as input the embeddings of a current observation from an MLLM backbone and predicts to keep or discard this observation, thus enabling an active participation of the MLLM in memory writing.

% We introduce two different ways to train the MemCtrl module: offline, supervised with expert demonstrations, and online, using reinforcement learning. We then provide extensive experiments on the EmbodiedBench interface, which encompasses ALFRED and Habitat benchmarks, showing the improved performance of MemCtrl with either training variant over full-context memory storage methods. We show consistent improvement in results across Gemma and Qwen models, highlighting the effectiveness of the memory head. 

\section{Limitations}
Our work has a few limitations. The supervised learning method requires expert demonstrations from a stronger model to understand which observations contribute to success, and the RL variants suffer from the inefficiencies that come with sparse reward structures. Future work could look into designing better reward functions that can capture the \textit{interesting-ness} of an observation \cite{interesting}. 
Furthermore, the benefits of MemCtrl degrade over short horizons, suggesting a limited need to train a memory head to filter observations in more basic settings. 
Another possible avenue we are excited to explore is to incorporate audio observations to increase the complexity of observations stored. Sim-to-real transfer of our work via real world experiments is also a potential extension.
% {
%     \small
%     \bibliographystyle{ieeenat_fullname}
%     \bibliography{main}
% }
\bibliography{main}

\appendix
\newpage

\begin{figure*}[t!]
    \centering
    \includegraphics[width=\linewidth]{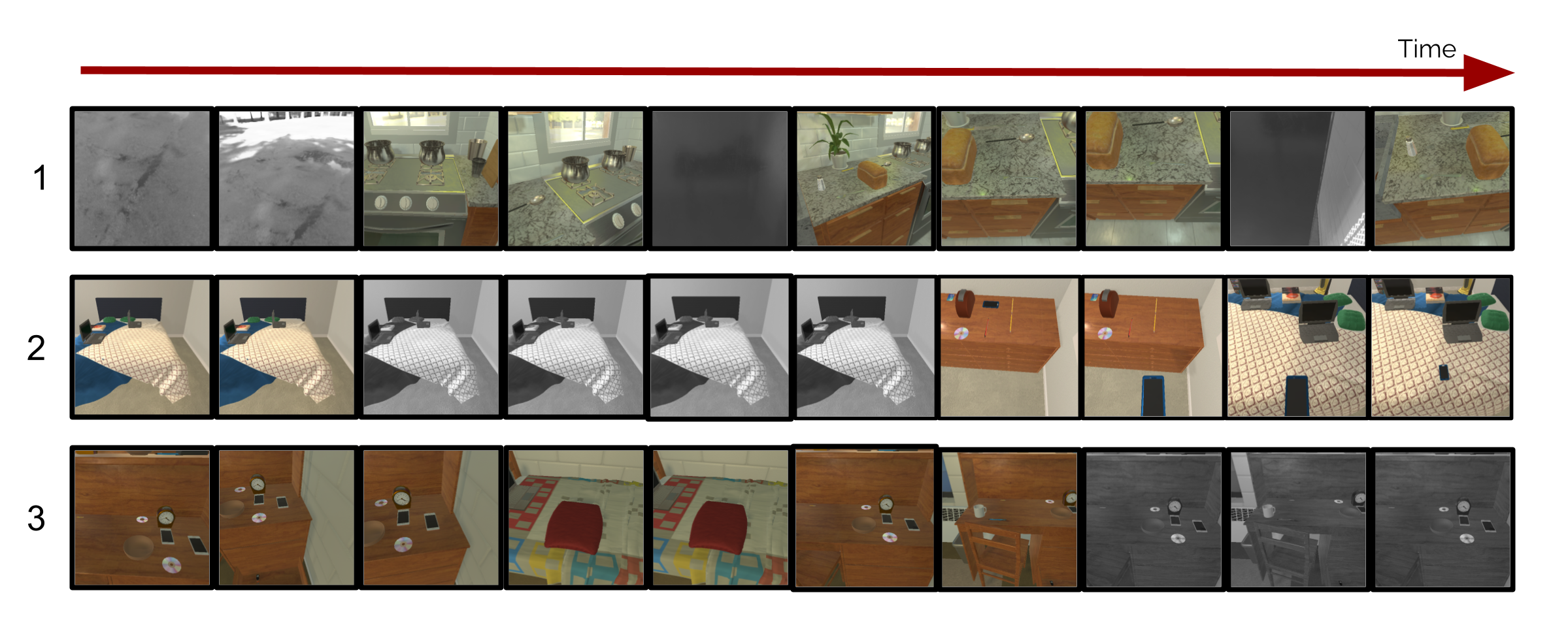}
    \caption{\textbf{Active Filtering}: The transferrable memory head performs active filtering, i.e., filters observations on the go. The grayed out images represent discarded ones. In sequence 1, notice MemCtrl filters out redundant images taken at odd angles while the agent looks around. In sequence 2, the agent makes a bunch of invalid actions in the middle of the sequence, before ultimately completing its task of placing a phone on the bed. These extra observations are filtered out as an outcome of our negative dense reward on invalid actions. In sequence 3, notice the repeating pattern towards the end, which starts to get filtered out by $\mu$. In each of these cases, the active involvement of the MLLM in filtering allows for better write-time memory control.}
    \label{fig:comparison}
\end{figure*}

\section{Algorithms}

\begin{algorithm}[h]
\begin{algorithmic}[1]
\Require  Labeled ground-truth expert answers $E = \{(y_i, \mathcal{O}^E_i)\}_{i=1}^K$, Memory Head $\mathcal{M}_{\mu}$, Action Head $\mathcal{M}_{a}$
\State \textit{Pretraining $\mu$}
\For{$i$ in $[1,K]$}
\State $\hat{p}_i = \mu(\mathcal{O}^E_i)$
\State Calculate loss $L(y_i, \hat{p}_i)$ with Equation \ref{eq:ce} 
\State Update $\mu$
\EndFor
\State \textit{Test-time with trained $\mu$}
\State $c \leftarrow F(\mathcal{C}, I)$ 
\State $a \leftarrow \mathcal{M}_{a}(\mathcal{O}_{c}, \mathcal{I}, c)$
\State $b \leftarrow \mathcal{M}_{\mu}(\mathcal{O}_{c}, \mathcal{I}, c)$
\If{$b = 1$}
    \State $\mathcal{C} \leftarrow \mathcal{C} \cup \{(\mathcal{O}_{c}, a)\}$
\EndIf
\end{algorithmic}
\caption{Training Memory Heads $\mu$ with Offline, Supervised Learning.}\label{alg:supervised_method}
\end{algorithm}

\begin{algorithm}[h]
\begin{algorithmic}[1]
\Require Current observation $\mathcal{O}_{c}$, Instruction $\mathcal{I}$, Total Context $\mathcal{C}$, Memory Head $\mathcal{M}_{\mu}$, Action Head $\mathcal{M}_{a}$
\State $c \leftarrow F(\mathcal{C}, I)$ 
\State $a \leftarrow \mathcal{M}_{a}(\mathcal{O}_{c}, \mathcal{I}, c)$
\State $b \leftarrow \mathcal{M}_{\mu}(\mathcal{O}_{c}, \mathcal{I}, c)$
\If{$b = 1$}
    \State $\mathcal{C} \leftarrow \mathcal{C} \cup \{(\mathcal{O}_{c}, a)\}$
\EndIf
\State Calculate reward with Equation \ref{eq:reward}
\State Update $\mu$
\end{algorithmic}
\caption{Training Memory Heads $\mu$ with Online, RL.}\label{alg:rl_method}
\end{algorithm}

\section{Experimental Details}

We train the memory head $\mu$ both via offline supervised learning with an expert, and online RL. In both cases, $\mu$ is an MLP initialized with $3$ layers to map the hidden MLLM dimension to a binary output.

\noindent\textbf{Expert, Offline Supervised}: Here, we first gather a dataset using an expert model that performs well on EmbodiedBench. We use GPT-4o for this, which has high success rates of $56.3\%$ and $59.0\%$ on EB-ALFRED and EB-Habitat respectively \cite{yang2025embodiedbench}. We gather observations, actions predicted, and episode success. We use this to create labels for our loss function, where if the action predicted was valid (\textit{last\_action\_success} variable in EmbodiedBench is true) or the episode was successful, we associate the image and it's visual state description (text) to a positive label, and negative otherwise. We make sure to balance out this dataset for optimal training.
We then train $\mu$ with the gathered labels as ground truth, and embeddings from the MLLM backbone (Qwen or Gemma) model taking in the image and visual scene description as input. We use a Binary Cross-Entropy Loss with the Adam optimizer. Our learning rate is $1e-3$.

\noindent\textbf{RL, Online}: Here, we train online using REINFORCE. At each timestep, we sample a binary action from the current policy, which in this case is whether to keep or discard the memory. We then compute the cumulative reward after executing the actions predicted by the action head, and update the policy.
To compute the cumulative reward, we use 1) a dense reward from the action head predicting a valid action, and 2) a sparse reward from episode success. Observe that both these rewards are not directly related to keeping or discarding memory, but are instead a result of the agent performing expected behavior to improve success. Modeling direct rewards to determine which memory to keep or discard is challenging, as they are tied to the nature of the task. For instance, an observation containing a white wall might not be useful for tasks involving picking up objects, but becomes necessary when it comes to answering questions about the environment. However, given enough training data, an agent can learn about the types of tasks being asked and corresponding memory fragments to keep in order to successfully complete them. This is analogous to a lifelong learning agent that must continually adapt to its surroundings to provide personalized assistance.

% \section{Memory Type: Complete vs Selective}

% \section{Comparison with RAG}

% A popular method 

\section{MLLM Prompts \& Decoder Modifications}

We modify the base prompt provided by EmbodiedBench with stricter constraints on choosing actions by adding the following:
\begin{quote}
\textit{\textbf{Important Rules}: \\
The action\_id must be picked from the available ids provided above. \\
Make sure the action\_id matches the corresponding action\_name. \\
A valid path is guaranteed to exist. If the image does not contain the required object for completing the task, you may have to navigate there.
}
\end{quote}

We also decode by action name instead of action id, and note that this leads to a significant increase in the number of valid actions taken while planning. For instance, Qwen outputs: 

\begin{quote}
\textit{{
visual\_state\_description: ``The scene ..."\\
reflection\_and\_reasoning: ``The user wants ... "\\
executable\_plan: \{$51$: ``pick up the hammer", ...\}
}}
\end{quote}
We note that the action of ``pick up the hammer'' exists and is valid, but it is linked to a different action\_id, $28$, causing the action decoding to fail on the base EmbodiedBench code. 

We speculate that this may be connected to prior literature showing that LLMs struggle with multiple choice selection \cite{xue2024strengthened, zheng2024largelanguagemodelsrobust}. The action selection of this work can be viewed as a multiple choice problem; therefore, some of the issues of selecting a natural language phrase with an ID can exist here as well. Furthermore, the action to ID mapping is at the beginning of the prompt, and there is research that shows that where information is positioned in the prompt affects reasoning ability \cite{cobbina-zhou-2025-show}. 
Another reason could be that smaller sized LLMs are better at generating descriptive language with a larger token counts, and lack the capacity for integer action mapping requiring consolidation towards a smaller token count \cite{kim2024palm}.

For this work, we modify the decoding function to map the \textit{action name} instead of the ID and notice improved performance, especially on the Qwen model.
Further work could potentially try moving the mapping information to different locations in the prompt. Another potential cause could be prompt length, where irrelevant information or even increased whitespace can degrade LLM accuracy, as shown in  \citet{du2025contextlengthhurtsllm}. Reducing the prompt to remove unnecessary whitespace or characters could potentially show improvement in outputting both the correct action name and the corresponding ID.

\section{Analyzing Memory}

The memory head enables a form of selective memory as an alternative to passing a complete history of observations. To highlight its effectiveness, we perform an ablation with complete memory, where none of the observations are discarded, and all of them are passed back to the MLLM, while maintaining a token horizon to prevent overflow. Table \ref{tab:selvscomp} presents results for the complete memory case on EB-Habitat and EB-ALFRED, in comparison to $\mu_{RL}$.

\begin{table*}[t!]
\centering
\resizebox{\linewidth}{!}{
\begin{tabular}{l|cccccc|cccccc}
\toprule
\textbf{Model} &
\multicolumn{6}{c|}{\textbf{EB-ALFRED}} &
\multicolumn{6}{c}{\textbf{EB-Habitat}} \\
 & \textbf{Avg} & \textbf{Base} & \textbf{Common} & \textbf{Complex} &
   \textbf{Spatial} & \textbf{Long} &
   \textbf{Avg} & \textbf{Base} & \textbf{Common} & \textbf{Complex} &
   \textbf{Spatial} & \textbf{Long} \\
\midrule
Qwen2.5-VL-7B-Ins & $4.7$ & $\underline{10}$ & $8$ & $6$ & $0$ & $2$ & $14.3$ & $32$ & $2$ & $26$ & $14$ & $2$ \\
Qwen2.5-VL-7B-Ins + $\mu_{\textbf{Complete}}$ & $7.8$ & $8$ & $7$ & $18$ & $1$ & $5$ & $8.8$ & $28$ & $0$ & $8$ & $8$ & $0$ \\
Qwen2.5-VL-7B-Ins + $\mu_{\textbf{RL}}$ & $\textbf{14.2}$ & $\underline{10}$ & $\underline{13}$ & $\underline{21}$ & $\underline{3}$ & $\underline{24}$ & $\textbf{22.2}$ & $\underline{37}$ & $\underline{3}$ & $\underline{37}$ & $\underline{16}$ & $\underline{18}$ \\
\bottomrule
\end{tabular}}
\caption{\textbf{Complete vs Selective}: We compare the performance of complete memory $\mu_\textbf{Complete}$ where all observations are passed as context to our best performing selective memory agent, $\mu_{\text{RL}}$. Note the improved performance of our selective memory agent, highlighting the importance of being picky about what to store in memory, especially when model capacity is limited.}
\label{tab:selvscomp}
\end{table*}

\begin{table*}[t]
\centering
\resizebox{\linewidth}{!}{
\begin{tabular}{l|cccccc|cccccc}
\toprule
& \multicolumn{6}{c|}{\textbf{EB-ALFRED}} 
& \multicolumn{6}{c}{\textbf{EB-Habitat}} \\
\textbf{Method} & \textbf{Avg} &
\textbf{Base} & \textbf{Common} & \textbf{Complex} & \textbf{Spatial} & \textbf{Long} &
\textbf{Avg} &
\textbf{Base} & \textbf{Common} & \textbf{Complex} & \textbf{Spatial} & \textbf{Long} \\
\midrule
\multicolumn{11}{l}{\textit{Memory Efficiency} $\mu_{\mathcal{E}}  (\%)$ $\downarrow$} \\
Qwen2.5 (Baseline, No Mem.)                
    & $N/A$ & $N/A$ & $N/A$ & $N/A$ & $N/A$  
    & $N/A$ & $N/A$ & $N/A$ & $N/A$ & $N/A$ & $N/A$ & $N/A$ \\
Qwen2.5 + $\mu_{\text{RL}}$                
    & $39.42$ & $\underline{35.6}$ & $42.8$ & $\underline{39.9}$ & $38.7$  
    & $40.1$ & $27.56$ & $39.2$ & $13.7$ & $15.7$ & $\underline{22.9}$ & $46.3$ \\
Qwen2.5 + $\mu_{\text{Expert}}$            
    & $\textbf{38.66}$ & $37.9$ & $\underline{40.1}$ & $45.6$ & $\underline{36.4}$ & $\underline{33.3}$  
    & $\textbf{26.38}$ & $\underline{37.2}$ & $\underline{10.2}$ & $\underline{14.8}$ & $36.5$ & $\underline{33.2}$ \\
Qwen2.5 + $\mu_{\text{Complete}}$          
    & $100$ & $100$ & $100$ & $100$ & $100$  
    & $100$ & $100$ & $100$ & $100$ & $100$ & $100$ & $100$ \\
\midrule

\multicolumn{11}{l}{\textit{Invalid Actions $\mathcal{I}$} $\downarrow$} \\
Qwen2.5 (Baseline, No Mem.)                
    & $3.50$ & $3.1$ & $2.9$ & $4.6$ & $3.8$  
    & $3.1$ & $3.0$ & $0.5$ & $5.2$ & $4.8$ & $2.7$ & $1.8$ \\
Qwen2.5 + $\mu_{\text{RL}}$                
    & $2.22$ & $\underline{2.0}$ & $1.8$ & $2.9$ & $2.1$  
    & $2.3$ & $1.36$ & $\underline{0.4}$ & $3.0$ & $2.4$ & $\underline{0.6}$ & $\underline{0.3}$ \\
Qwen2.5 + $\mu_{\text{Expert}}$            
    & $\textbf{2.10}$ & $2.7$ & $\underline{1.5}$ & $\underline{2.4}$ & $\underline{1.8}$  
    & $\underline{2.1}$ & $\textbf{1.02}$ & $0.6$ & $\underline{2.2}$ & $\underline{1.3}$ & $\underline{0.6}$ & $0.4$ \\
Qwen2.5 + $\mu_{\text{Complete}}$          
    & $3.10$ & $2.7$ & $4.2$ & $3.3$ & $2.3$  
    & $3.0$ & $2.12$ & $1.5$ & $4.6$ & $2.8$ & $0.8$ & $0.9$\\
\bottomrule
\end{tabular}}
\caption{
\textbf{Statistics}: Memory efficiency ($\uparrow$) and invalid actions ($\downarrow$) across all five splits for EB-Habitat and EB-Alfred. $\mu_{\mathcal{E}}$ for the expert memory head is slightly better on average, but is much worse on ALFRED. This can be attributed to tasks in ALFRED being slightly harder than Habitat. $\mu$ augmented Qwen models also make lesser number of invalid actions per episode. Overall, adding a memory head shows significant improvement over no memory and complete memory baselines.
}
\label{tab:eff}
\end{table*}

\subsection{Statistics}
Table \ref{tab:eff} highlights statistics gathered across all $5$ splits on EB-Habitat and EB-Alfred for Qwen and the memory augmentations. We measure the following across $20$ randomly chosen episodes:

\noindent\textbf{Memory Efficiency $\mathcal{E}$}:
To determine the effectiveness of our approach, we compute the memory efficiency per episode as, 
\[
\mathcal{E} = 1 - \frac{\text{Number of Memories Kept}}{\text{Total Steps Taken}} 
\]

This gives us the fraction of memories that were stored in the memory bank per episode. When all the memories have been store, $\mathcal{E}$ resolves to $0$, meaning the agent was inefficient in its memory management. 

\noindent \textbf{Invalid Actions $\mathcal{I}$}: This is the average number of times that the MLLM responds with an invalid action for execution. For instance, the MLLM asks the agent to \textit{`pick up a spoon'}, but there is no spoon visible in the observation.

\noindent \textbf{Inferences}: The table highlights the improved memory efficiency of both our models. In our main results table, we noted the improved performance of the online RL memory head on Qwen. By multiplying values from that table with the efficiency values here, we get a \emph{weighted efficiency} score that aims to capture both success and frugality of memory usage. This can be written as,
\[
\mathcal{W}^{(m)_{b}}
= \text{Succ.}^{(m)} \cdot \left(1 - \frac{\mu_{\mathcal{E}}^{(m)}}{100}\right),
\]
where $\text{Succ}^{(m)}$ is the task success (in \%) of method $m$ on benchmark $b \in \{\text{Alfred, Habitat}\}$, and $\mu_{\mathcal{E}}^{(m)}$ is the corresponding memory efficiency. We then aggregate this into a single score per method as,
\[
{\mathcal{W}}^{(m)}
= \frac{\mathcal{W}^{(m)}_{Alf.} + \mathcal{W}^{(m)}_{Hab.}}{2}
\]

Substituting the values from Table 1 in the main text and Table 2 in this supplementary, we get 
$\mathcal{W}^{\text{Exp.}} = 12.13 $ and $\mathcal{W}^{\text{RL}} = 10.95$, meaning that the expert has slightly better overall weighted efficiency.

We also note that the invalid actions $\mathcal{I}$ are significantly lower on both our memory heads when compared to the baseline and complete memory approaches. This further highlights the effectiveness of selective memory where $\mu$ actively decides to populate the context for the MLLM on the go.
% \noindent \textbf{Steps $\mathcal{S}$}: We also note the average number of steps taken by the agent across episodes. 

Across the 3 sequences in Figure \ref{fig:comparison}, note how our approach actively filters redundant and repeated observations. A trained memory head can be transferred to novel settings, learning to distinguish between interesting and non-interesting observational data \cite{interesting}.

\end{document}

%% file: sec/0_abstract.tex
\begin{abstract}

Foundation models rely on in-context learning for personalized decision making. The limited size of this context window necessitates memory compression and retrieval systems like RAG. These systems however often treat memory as large offline storage spaces, which is unfavorable for embodied agents that are expected to operate under strict memory and compute constraints, online.
In this work, we propose MemCtrl, a novel framework that uses Multimodal Large Language Models (MLLMs) for pruning memory online. MemCtrl augments MLLMs with a \textit{trainable} memory head $\mu$ that acts as a gate to determine which observations or reflections to retain, update, or discard during exploration. We evaluate with training two types of $\mu$, 1) via an offline expert, and 2) via online RL, and observe significant improvement in overall embodied task completion ability on $\mu$-augmented MLLMs. In particular, on augmenting two low performing MLLMs with MemCtrl on multiple subsets of the EmbodiedBench benchmark, we observe that $\mu$-augmented MLLMs show an improvement of around $16\%$ on average, with over $20\%$ on specific instruction subsets.
Finally, we present an qualitative analysis on the memory fragments collected by $\mu$, noting the superior performance of $\mu$ augmented MLLMs on long and complex instruction types.
\end{abstract}